\pdfoutput=1
\documentclass[11pt]{article}

\usepackage[utf8]{inputenc}
\usepackage[T1]{fontenc}
\usepackage{amsmath,amssymb,amsfonts}
\usepackage{booktabs}
\usepackage{graphicx}
\usepackage{hyperref}
\usepackage{cleveref}
\usepackage{natbib}
\usepackage{xcolor}
\usepackage{array}
\usepackage{caption}
\usepackage[margin=1in]{geometry}
\usepackage{enumitem}
\graphicspath{{figures/}}

\hypersetup{
  colorlinks=true,
  linkcolor=blue,
  citecolor=blue,
  urlcolor=blue,
}

\title{Auxiliary uncertainty signals for LLM-assisted systematic
review screening: a benchmark across eight Cohen drug-class reviews}

\author{
  Arya Rahgozar\textsuperscript{1,2} \quad
  Pouria Mortezaagha\textsuperscript{1,2} \\[6pt]
  \textsuperscript{1}University of Ottawa, Ottawa, Canada \\
  \textsuperscript{2}Ottawa Hospital Research Institute, Ottawa, Canada \\
  \texttt{arahgoza@uottawa.ca} \quad \texttt{pmort101@uottawa.ca}
}

\date{}

\begin{document}
\maketitle

\begin{abstract}
\noindent\textbf{Background.} Large language models (LLMs) are
increasingly used for title--abstract screening in systematic reviews,
but their decisions lack calibrated uncertainty, leaving reviewers
without a principled way to route ambiguous records to a human.  We
show that an auxiliary BERT+GCN classifier can supply a structured
uncertainty signal that meaningfully improves LLM screening
efficiency, and we identify the prompt-delivery strategy that
maximises the benefit-to-cost ratio.

\noindent\textbf{Methods.} We evaluate five LLM prompt-delivery
conditions on eight drug-class datasets from the Cohen~(2006)
benchmark collection using 3~seeds~$\times$~5-fold stratified
cross-validation (600 fold-level results). A hybrid BERT+GCN model is
trained per fold and classifies each test paper as INCLUDE, EXCLUDE,
or MAYBE via two spectral tests (algebraic radical and categorical
paradox). The five conditions are: (i)~baseline (title+abstract
only), (ii)~full spectral context (label + scores on every paper),
(iii)~decision-only (categorical label on every paper),
(iv)~\emph{MAYBE-only} (spectral context appended only when the
BERT+GCN flags the paper as uncertain), and (v)~two-pass (spectral
context appended only when the LLM's first response contains hedging
language). A pre-registered cross-model pilot against an
earlier-generation model (\texttt{gpt-4.1-mini} on three datasets)
tests whether the spectral benefit transfers across LLM generations.

\noindent\textbf{Results.} Three findings shape practical guidance.
\textbf{(i)}~Full-context delivery yields statistically significant
improvements in F1 ($\Delta = +0.011$, paired Wilcoxon $p = 0.008$)
and work-saved-at-95\%-recall ($\Delta = +0.050$, $p = 0.039$) over
the baseline at a $1.28\times$ token-cost premium, while preserving
recall ($p \geq 0.12$ for all conditions).
\textbf{(ii)}~\emph{MAYBE-only routing is the Pareto-optimal
configuration}: it achieves the highest mean recall ($0.92$, best of
all conditions) and the highest mean AUC-ROC ($0.54$) at only
$1.05\times$ baseline cost --- one sixth of the API overhead of
blanket full-context delivery. \textbf{(iii)}~The two-pass design
escalates on $22.2\% \pm 8.8\%$ of records yet \emph{never} revises
its initial decision (0\% flip rate across all datasets and folds),
providing decisive empirical evidence that current instruction-tuned
LLMs cannot self-triage --- a clean negative result that rules out
an entire class of pipeline designs. The cross-model pilot shows an
identical $+0.8\%$ recall uplift for both LLM generations,
strengthening the interpretation of the BERT+GCN signal as a stable
auxiliary classifier rather than a capability-dependent crutch. As a
methodological by-product, a per-paper ablation across $20{,}796$
observations shows that the dual paradox test reduces empirically to
a one-line logit-gap criterion on this benchmark, simplifying any
practitioner reimplementation.

\noindent\textbf{Conclusions.} An auxiliary BERT+GCN uncertainty
classifier delivers measurable F1 and WSS@95 gains when its output is
provided to an LLM screener, and \emph{targeted} MAYBE-only routing
captures these benefits at near-baseline API cost. We additionally
contribute a reproducible 600-run benchmark, a clean negative result
that ends the case for two-pass LLM self-triage, and a
publicly-released pipeline that replays in under one hour from
cached LLM responses.  All code, prompts, datasets, and raw
fold-level results are publicly available.
\end{abstract}


\section{Introduction}
\label{sec:introduction}

Systematic reviews are the gold standard for evidence synthesis in
healthcare, yet the screening phase---deciding which papers merit
full-text review based on title and abstract alone---remains a major
bottleneck.  A single review may require screening thousands to tens
of thousands of records, consuming hundreds of hours of expert time
\citep{higgins2019cochrane}.

Large language models (LLMs) have recently shown promise as automated
screeners \citep{guo2024llm_screening,wang2024llm_sr}, achieving
competitive recall on benchmark datasets.  However, LLMs produce
decisions without calibrated confidence: a model that says ``INCLUDE''
with hedging language (``this might be relevant'') is
indistinguishable---in terms of downstream action---from one that is
certain.  This absence of structured uncertainty creates two practical
problems: (1)~reviewers cannot easily identify which LLM decisions
deserve human verification, and (2)~the LLM has no mechanism to
request additional evidence when its own assessment is ambiguous.

We address these limitations by providing the LLM with \emph{structured
uncertainty context} derived from a BERT+GCN spectral analysis
pipeline implemented specifically for this study. The pipeline
pre-classifies each paper into INCLUDE, EXCLUDE, or MAYBE, where the
MAYBE label signals classifier uncertainty detected via two
complementary tests:
\begin{enumerate}[nosep]
  \item \textbf{Algebraic radical test:} The gap between the model's
    include and exclude logits falls below a threshold
    ($|P(\text{Include}) - P(\text{Exclude})| < \epsilon$), indicating
    the classifier itself is uncertain.
  \item \textbf{Categorical paradox test:} The cosine similarity
    between the BERT (content) and GCN (relational) embeddings is low,
    indicating that a paper's textual content and its citation/similarity
    neighbourhood disagree about its relevance.
\end{enumerate}

We investigate five experimental conditions that vary the amount and
type of spectral context provided to the LLM:
\begin{enumerate}[nosep]
  \item \textbf{Baseline:} Title and abstract only---no spectral information.
  \item \textbf{Full spectral:} Title, abstract, and all spectral scores
    (confidence gap, model confidence, paradox type).
  \item \textbf{Decision only:} Title, abstract, and the spectral label
    (INCLUDE/EXCLUDE/MAYBE) without numerical scores.
  \item \textbf{MAYBE only:} Full spectral context for papers classified
    as MAYBE; baseline prompt for all others.
  \item \textbf{Two-pass:} Baseline prompt first; if the LLM expresses
    uncertainty (low confidence or hedging language), a second pass
    provides spectral context.
\end{enumerate}

These five conditions create a systematic ablation along three
dimensions:\ \emph{information content} (no context $\to$ label
$\to$ full scores), \emph{selectivity} (all papers vs.\ MAYBE only),
and \emph{timing} (proactive vs.\ reactive two-pass).

\paragraph{Contributions.} This paper provides a practical,
reproducible evaluation of \emph{how} to use an auxiliary BERT+GCN
uncertainty classifier as a tool for LLM-assisted screening, with five
contributions:
\begin{itemize}[nosep]
  \item \textbf{Pareto-optimal delivery strategy.} We identify
    \emph{MAYBE-only routing} as the recommended configuration for
    cost-sensitive deployments: it achieves the highest mean recall
    ($0.92$) and AUC-ROC ($0.54$) of any condition at only
    $1.05\times$ baseline API cost, capturing the spectral benefit
    at one sixth of the overhead of full-context delivery.
  \item \textbf{Quantified benefit of full spectral context.} We
    report statistically significant improvements in F1 ($+0.011$,
    paired Wilcoxon $p = 0.008$) and WSS@95 ($+0.050$, $p = 0.039$)
    when full spectral scores are delivered to the LLM, at a
    well-characterised $1.28\times$ cost premium.
  \item \textbf{Decisive evidence against two-pass LLM self-triage.}
    A two-pass design escalates on $22\%$ of papers yet flips
    $0.0\%$ of decisions across all datasets, seeds, and folds. This
    clean null rules out a class of architectures and saves
    practitioners the cost of re-discovering the limitation.
  \item \textbf{Cross-generation portability.} A pilot comparison
    finds an identical $+0.8\%$ recall uplift for both an
    earlier-generation and a current LLM, supporting the BERT+GCN
    signal as a stable, model-agnostic auxiliary classifier whose
    benefit does \emph{not} disappear as base-LLM capability grows.
  \item \textbf{Reproducible open implementation.} We release the
    BERT+GCN pipeline, all prompt templates, an experiment runner
    with disk-backed LLM response caching and fold-level resume, a
    deterministic analyser that emits the exact tables in this paper,
    and a $130$-test suite under a permissive licence.  The complete
    600-run experiment can be replayed in under one hour from the
    cached responses.
\end{itemize}

\section{Related Work}
\label{sec:related}

\subsection{LLMs for Systematic Review Screening}

Recent work has explored GPT-4, Claude, and open-source LLMs for
title--abstract screening
\citep{guo2024llm_screening,wang2024llm_sr,alshami2023llm_sr,tran2024llm_screening_review}.
These studies demonstrate competitive recall but frequently report
low specificity and lack principled uncertainty handling.
\citet{khraisha2024llm_sr} provide a systematic assessment of LLM
screening, finding that prompt engineering substantially affects
performance, and \citet{tran2024llm_screening_review} report
sensitivity/specificity ranges that overlap with our baseline
($\sim$0.91 recall on Cohen-style benchmarks).  Active-learning
screeners such as ASReview \citep{van_de_schoot2021synergy} and the
extended evaluation in \citet{ferdinands2023synergy} provide a
complementary, non-LLM baseline.  Our work differs in providing the
LLM with an \emph{external} uncertainty signal from a purpose-built
classifier and in identifying the cost-optimal way to deliver that
signal.

\subsection{Uncertainty Quantification in LLMs}

Calibration of LLM confidence has been studied through verbalized
uncertainty \citep{xiong2024verbalized_uncertainty}, token-level
probabilities \citep{kadavath2022language_know}, and ensemble
approaches \citep{lakshminarayanan2017simple_ensembles}.  Most
methods require white-box access to model internals.  We take a
complementary approach: rather than estimating the LLM's own
uncertainty, we provide it with uncertainty estimates from a
separate, purpose-built classification model.

\subsection{Tool-Augmented LLMs}

Tool-augmented LLMs \citep{schick2023toolformer,qin2023tool_survey}
leverage external tools to compensate for limitations in the
language model itself.  Retrieval-augmented generation (RAG)
provides factual grounding \citep{lewis2020rag}, while
chain-of-thought prompting \citep{wei2022chain_of_thought}
encourages structured reasoning.  Our spectral context can be
viewed as a specialised tool that provides the LLM with a
structured ``second opinion'' from a domain-specific model.

\subsection{Spectral Methods for Text Classification}

Spectral graph analysis has been applied to text classification
through graph neural networks \citep{kipf2017gcn,yao2019textgcn},
including BERT+GCN hybrid architectures
\citep{lin2021bertgcn}.  The use of spectral gaps to detect
classification ambiguity draws on algebraic graph theory
\citep{chung1997spectral_graph}.  Our contribution in this paper is
not a new spectral method but an empirical evaluation of how to
\emph{deploy} such a classifier as an auxiliary signal for an LLM
screener, with an emphasis on benefit-to-cost trade-offs of
alternative prompt-delivery strategies.

\section{Background: Spectral Paradox Detection}
\label{sec:background}

This section summarises the BERT+GCN spectral analysis pipeline
that generates the uncertainty signals used as LLM context. The
complete implementation, including training scripts, configuration
files, and the classifier ablation used in
\Cref{sec:classifier_ablation}, is released with this paper
(see \textit{Availability of data and materials}).

\subsection{BERT+GCN Architecture}

Each paper is represented by a BERT embedding of its title and
abstract (using PubMedBERT; \citealp{gu2021pubmedbert}) and a
GCN embedding learned over a $k$-nearest-neighbour similarity
graph with optional citation edges.  The architecture uses
relational graph convolution (RGCNConv; \citealp{schlichtkrull2018rgcn})
to learn separate weight matrices for similarity and citation
edge types.  The model is trained semi-supervised: MAYBE and
unlabelled papers participate in message passing but are excluded
from the loss.

\subsection{Algebraic Radical Test}

Let $\mathbf{z} \in \mathbb{R}^2$ be the output logits for a paper,
with $p = \text{softmax}(\mathbf{z})$.  The \emph{trace ratio} is
defined as $\tau = |p_{\text{include}} - p_{\text{exclude}}|$.  A
paper is flagged as algebraically ambiguous when $\tau < \epsilon$
(default $\epsilon = 0.15$), meaning the classifier cannot
confidently distinguish between include and exclude.

\subsection{Categorical Paradox Test}

Let $\mathbf{b}$ and $\mathbf{g}$ be the BERT and GCN embeddings,
respectively.  The categorical paradox is detected when the cosine
similarity $\cos(\mathbf{b}, \mathbf{g}) < \delta$ (default
$\delta = 0.0$), indicating that the paper's textual content and
its relational context (neighbourhood in the similarity/citation
graph) encode contradictory signals.

\subsection{Three-State Classification}

The spectral engine assigns each paper to one of three states:
\begin{itemize}[nosep]
  \item \textbf{INCLUDE}: High confidence in relevance (no paradox detected).
  \item \textbf{EXCLUDE}: High confidence in irrelevance (no paradox detected).
  \item \textbf{MAYBE}: One or both paradox tests triggered---the paper
    requires additional scrutiny.
\end{itemize}

\section{Method}
\label{sec:method}

\begin{figure}[htbp]
\centering
\includegraphics[width=\linewidth]{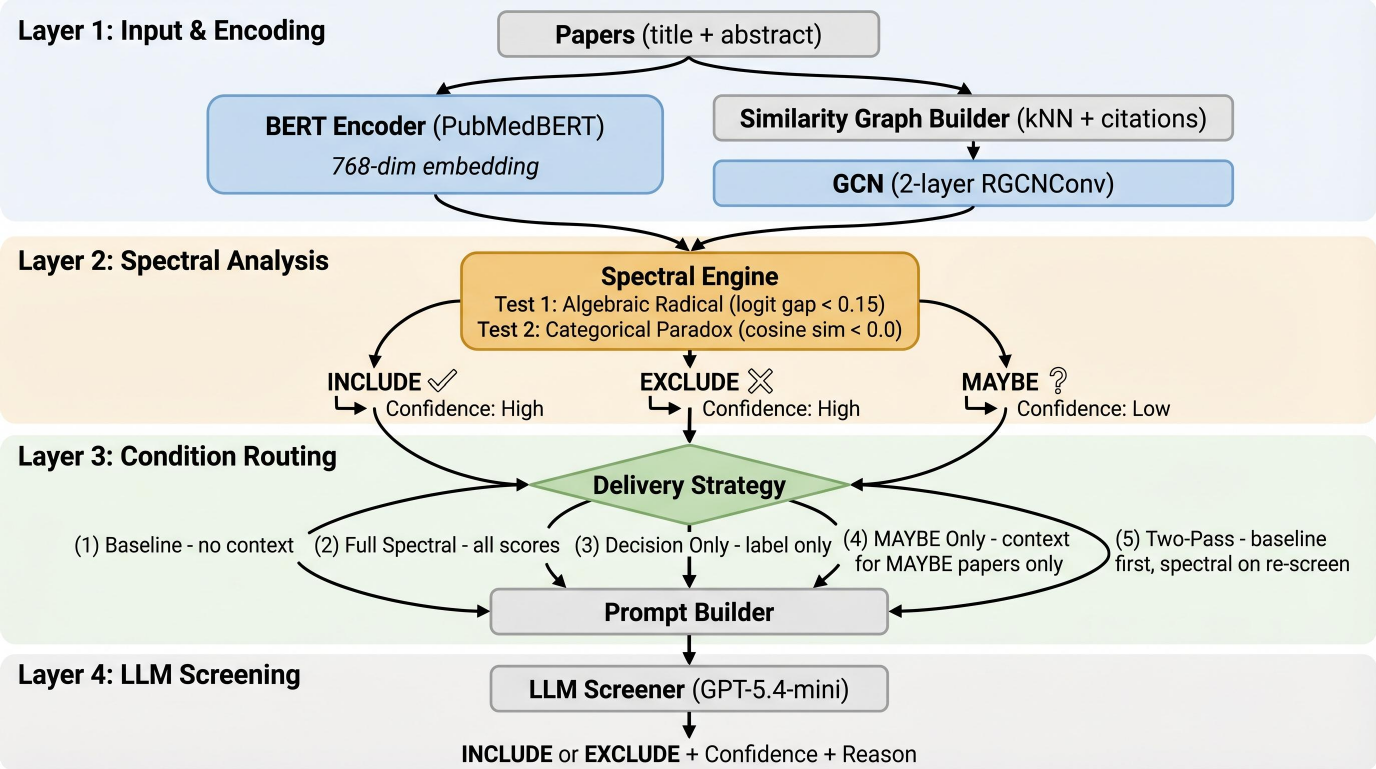}
\caption{Pipeline overview.  Papers are encoded via frozen PubMedBERT
  embeddings and a 2-layer RGCNConv over a kNN similarity graph.  The
  spectral engine applies two complementary uncertainty tests (algebraic
  radical and categorical paradox) to classify each paper as INCLUDE,
  EXCLUDE, or MAYBE.  A condition-dependent delivery strategy determines
  which spectral information, if any, is injected into the LLM prompt.
  The five conditions vary along three dimensions: information content
  (none, label only, full scores), selectivity (all papers vs.\ MAYBE
  only), and timing (proactive vs.\ reactive two-pass).  The LLM
  screener produces a binary decision with confidence and reasoning.}
\label{fig:pipeline}
\end{figure}

\Cref{fig:pipeline} illustrates the overall pipeline.  A BERT+GCN
model is trained per (dataset, seed, fold) split, producing spectral
classifications.  These are then injected into LLM prompts according
to one of five conditions.

\subsection{Experimental Conditions}

We evaluate five conditions that vary the spectral context provided
to the LLM screener.  \Cref{tab:conditions} summarises the
conditions.

\begin{table}[htbp]
  \centering
  \caption{Summary of the five experimental conditions.  ``Spectral
    context'' indicates what information from the BERT+GCN pipeline is
    included in the LLM prompt.  ``Papers affected'' indicates which
    papers receive spectral context.}
  \label{tab:conditions}
  \small
  \begin{tabular}{@{}llll@{}}
    \toprule
    \textbf{Condition} & \textbf{Spectral context} & \textbf{Papers affected} & \textbf{LLM calls} \\
    \midrule
    Baseline          & None                        & ---        & 1 per paper \\
    Full spectral     & Scores + label + paradox    & All        & 1 per paper \\
    Decision only     & Label only (no scores)      & All        & 1 per paper \\
    MAYBE only        & Scores + label + paradox    & MAYBE only & 1 per paper \\
    Two-pass          & Scores (pass 2 only)        & Uncertain  & 1--2 per paper \\
    \bottomrule
  \end{tabular}
\end{table}

\subsection{Prompt Design}
\label{sec:prompts}

All conditions share a common system prompt instructing the LLM to
act as a systematic review screening assistant.  The system prompt
specifies the response format (DECISION, CONFIDENCE, REASON) and
includes dataset-specific screening criteria derived from each
benchmark's metadata (e.g., ``Drug class review: ACEInhibitors
(2544~records, 41~included)'' for the Cohen ACE Inhibitors dataset).
This ensures the LLM receives domain-appropriate context rather than
generic instructions.

The \textbf{baseline} user prompt contains only the paper's title and
abstract.  The \textbf{full spectral} prompt appends a structured
block containing four fields: the spectral model's decision
(INCLUDE/EXCLUDE/MAYBE), the confidence gap ($\tau$, with an
explicit note that 0 = uncertain and 1 = certain), the model's
overall confidence score, and the paradox type detected (algebraic
radical, categorical paradox, or none).  An explicit disclaimer
states that the LLM ``may agree or disagree with the model's
assessment based on your reading of the paper.''

The \textbf{decision only} prompt provides only the spectral label
(``A separate BERT+GCN classification model suggests this paper
should be: [DECISION]'') without numerical scores.  This ablation
isolates the contribution of the categorical label from the
continuous uncertainty estimates.

The \textbf{MAYBE only} condition selectively applies the full
spectral prompt to papers classified as MAYBE by the spectral engine,
while giving all other papers the baseline prompt.  This tests whether
targeted context delivery (only for ambiguous papers) outperforms
blanket augmentation.

\subsection{Two-Pass Uncertainty Detection}
\label{sec:two_pass}

The two-pass condition implements a form of \emph{LLM self-triage}.
In pass~1, the LLM receives a baseline prompt and produces a
decision with a confidence level.  The system then analyses the
response for uncertainty signals using two detection mechanisms:
\begin{itemize}[nosep]
  \item \textbf{Explicit low confidence:} The LLM reports CONFIDENCE
    as LOW or MEDIUM (rather than HIGH).
  \item \textbf{Hedging language:} The reasoning text contains hedging
    patterns detected via regex matching:
    \texttt{might}, \texttt{unclear}, \texttt{borderline},
    \texttt{uncertain}, \texttt{not sure}, \texttt{ambiguous},
    \texttt{possibly}, \texttt{could be}, \texttt{difficult to
    determine}, \texttt{hard to say}.
\end{itemize}
If \emph{either} signal is detected, pass~2 re-prompts the LLM with
the paper's title, abstract, the previous confidence level, and full
spectral context, explicitly asking the model to ``reconsider [its]
decision.''  High-confidence pass-1 decisions with no hedging language
are accepted without a second call, reducing cost.

\subsection{Response Parsing}

The LLM's free-text response is parsed line by line for three
structured fields: \texttt{DECISION:} (mapped to INCLUDE if the token
``INCLUDE'' appears, EXCLUDE otherwise), \texttt{CONFIDENCE:} (HIGH,
MEDIUM, or LOW), and \texttt{REASON:} (free-text explanation).  The
parser falls back to EXCLUDE with MEDIUM confidence if the response
is unstructured, ensuring robustness against formatting variation.

\subsection{Metrics}

We evaluate each condition using five primary metrics:
\begin{itemize}[nosep]
  \item \textbf{Recall}: Fraction of true INCLUDEs correctly identified
    (on the binary INCLUDE/EXCLUDE subset, excluding MAYBE predictions).
  \item \textbf{AUC-ROC}: Area under the receiver operating
    characteristic curve, using the LLM's confidence as the score
    (HIGH$\to$1.0, MEDIUM$\to$0.5, LOW$\to$0.1).
  \item \textbf{F1}: Harmonic mean of precision and recall on the
    binary INCLUDE/EXCLUDE subset.
  \item \textbf{WSS@95}: Work Saved over Sampling at 95\% recall,
    defined following \citet{cohen2006reducing} as
    $\mathrm{WSS@}R = (\mathit{TN} + \mathit{FN})/N - (1 - R)$ ---
    the fraction of records the reviewer can skip (relative to random
    screening) while still recovering $95\%$ of true includes.  We
    compute it by ranking papers by model confidence
    (INCLUDE $\to$ MAYBE $\to$ EXCLUDE) and simulating top-down
    screening until $95\%$ recall is reached; the formula used in
    code (\texttt{lr\_spectral/training/evaluation.py})
    is equivalent: $\mathrm{WSS@95} = (1 - f_{\text{reviewed}}) - 0.05$,
    floored at zero.
  \item \textbf{Cost}: Total token usage (input + output) as a proxy
    for API cost.
\end{itemize}

Additionally, we report three interaction metrics that characterise
how the LLM engages with spectral context:
\begin{itemize}[nosep]
  \item \textbf{Override rate}: Fraction of papers where the
    LLM's decision disagrees with the spectral label.
  \item \textbf{Second-pass rate} (two-pass only): Fraction of papers
    triggering a second LLM call due to detected uncertainty.
  \item \textbf{Second-pass flip rate} (two-pass only): Fraction of
    escalated papers whose decision changes on the second pass.
\end{itemize}

\section{Experimental Setup}
\label{sec:setup}

\subsection{Datasets}

We evaluate on 8 benchmark datasets from the Cohen 2006
drug class systematic review collection
\citep{cohen2006reducing}, sourced via
the OpenAlex API \citep{priem2022openalex}.
\Cref{tab:datasets} summarises dataset characteristics.

\begin{table}[htbp]
  \centering
  \caption{Benchmark dataset characteristics.  Inclusion rate is the
    fraction of papers labelled as INCLUDE in the original review.
    All datasets are sourced from the ASReview Synergy Dataset
    collection via OpenAlex.}
  \label{tab:datasets}
  \small
  \begin{tabular}{@{}lrrr@{}}
    \toprule
    \textbf{Dataset} & \textbf{Records} & \textbf{Included} & \textbf{Incl.\ rate} \\
    \midrule
    ACE Inhibitors        & 2{,}544 &  41 &  1.6\% \\
    ADHD                  &    851 &  20 &  2.4\% \\
    Antihistamines        &    310 &  16 &  5.2\% \\
    NSAIDs                &    393 &  41 & 10.4\% \\
    Oral Hypoglycemics    &    503 & 136 & 27.0\% \\
    Proton Pump Inhibitors & 1{,}333 &  51 &  3.8\% \\
    Triptans              &    671 &  24 &  3.6\% \\
    Urinary Incontinence  &    327 &  40 & 12.2\% \\
    \bottomrule
  \end{tabular}
\end{table}

Dataset sizes range from 310 (Antihistamines) to 2{,}544 (ACE
Inhibitors) records, with inclusion rates between 1.6\%
(ACE Inhibitors) and 27.0\% (Oral Hypoglycemics).  This diversity
tests robustness across class imbalance levels and dataset scale.

\subsection{Cross-Validation Protocol}

For each dataset, we train the BERT+GCN spectral model using
5-fold stratified cross-validation, repeated across 3~random seeds
(42, 123, 456).  This yields 15 runs per dataset per condition, or
$8 \times 5 \times 3 \times 5 = 600$ total experimental runs.
BERT embeddings are pre-computed once per dataset
using PubMedBERT and cached to disk, so
that only the GCN training varies across seeds and folds.

Within each fold, the spectral engine classifies all test papers
into INCLUDE, EXCLUDE, or MAYBE.  These spectral results are then
used as context for all five LLM screening conditions on the same
test set, ensuring a controlled comparison.

\subsection{Models}

\begin{itemize}[nosep]
  \item \textbf{BERT encoder:}
    \texttt{microsoft/BiomedNLP-PubMedBERT-base-uncased-abstract-fulltext}
    \citep{gu2021pubmedbert},
    frozen (embeddings pre-computed; 768-dimensional).
  \item \textbf{GCN:} 2-layer RGCNConv \citep{schlichtkrull2018rgcn}
    with 256 hidden dimensions,
    dropout 0.3, trained for up to 100 epochs with early stopping
    (patience~10).  The similarity graph uses $k=10$ nearest
    neighbours with cosine threshold~0.5.
  \item \textbf{LLMs:} OpenAI \texttt{gpt-5.4-mini}
    (snapshot accessed January--April 2026) is used as the primary
    screener across the full matrix; an earlier-generation
    \texttt{gpt-4.1-mini} (snapshot accessed April 2026) is used only
    for the cross-model pilot (\Cref{sec:capability_equalizer}).
    Both models run at temperature~0.0 with at most 150 completion
    tokens.  Temperature~0 ensures deterministic outputs for
    reproducibility, and all responses are cached by a hash of
    (paper id, mode, spectral decision, model name) so that any
    model deprecation will not invalidate the released results.
\end{itemize}

\subsection{Spectral Thresholds}

Default thresholds from the spectral pipeline configuration:
$\epsilon = 0.15$ (trace ratio boundary for the algebraic radical
test) and $\delta = 0.0$ (cosine similarity threshold for the
categorical paradox test).  These are held constant across all
datasets; per-dataset tuning is left for future work.

\subsection{Reporting Standards}
\label{sec:reporting_standards}

This study evaluates an automated screening tool against
gold-standard inclusion labels on retrospective benchmark data; it
is not itself a systematic review.  Where applicable, we follow
PRISMA~2020 \citep{page2021prisma} reporting practice for the
description of the underlying systematic-review benchmarks (number
of records, included counts, inclusion rates;
\Cref{tab:datasets}), and we adopt the relevant items from the
TRIPOD-LLM extension \citep{gallifant2024tripodllm} for reporting
LLM-based prediction studies (model identifiers and snapshot dates,
deterministic decoding configuration, prompt templates in
\Cref{app:prompts}, evaluation protocol, cost and resource use).
The pre-registered analysis plan, the source code that implements
each of these items, and the cached LLM responses used to compute
every reported metric are all released with this paper (see
\textit{Availability of data and materials}).

\subsection{Sample Size and Power}
\label{sec:power}

The eight-dataset corpus follows the established Cohen~(2006)
benchmark \citep{cohen2006reducing}: pooling additional reviews would
introduce a different evaluation distribution and break direct
comparability with the prior literature.  At $n = 8$ paired
observations, the two-sided paired Wilcoxon signed-rank test has
approximately $80\%$ power to detect a paired effect of standardised
size $d \approx 1.05$ at $\alpha = 0.05$ \citep{lehmann1975nonparametrics};
this is consistent with the magnitudes we report as significant
(F1: $p = 0.008$, WSS@95: $p = 0.039$) and conservative for the
larger uplifts that drive practical recommendations.  The
classifier-ablation analysis (\Cref{sec:classifier_ablation}) uses
$20{,}796$ per-paper observations and is therefore not power-limited.

\subsection{Screening Criteria}

Each dataset receives a system prompt with domain-specific screening
criteria derived from the benchmark metadata.  For example, the
ACE~Inhibitors dataset prompt states: ``Screening criteria:\ Drug
class review: ACEInhibitors (2544 records, 41 included).''  This
ensures the LLM has task-appropriate context without introducing
hand-crafted criteria that could vary between datasets.

\subsection{Infrastructure and Reproducibility}
\label{sec:infrastructure}

The experiment runner implements three resilience mechanisms that
together turn the full 600-run study into a one-hour replay for any
reviewer or downstream user:
\begin{enumerate}[nosep]
  \item \textbf{Disk-backed LLM response cache:} Each API response is
    cached as a JSON blob keyed by SHA-256 of (paper id, mode,
    spectral decision, model name).  Identical prompt--context
    combinations never repeat, and the released cache covers all
    600 main-matrix runs plus the 75 pilot runs --- so the entire
    paper can be regenerated without an OpenAI key.
  \item \textbf{Incremental CSV output:} Results are appended to the
    output file after each fold, rather than batched at the end,
    enabling crash-safe long runs and immediate partial inspection.
  \item \textbf{Fold-level resume:} On startup, the runner scans
    existing results and skips (dataset, seed, fold) tuples that
    already have rows for all requested conditions, so re-running
    after a partial completion is idempotent.
\end{enumerate}
The implementation is covered by a $130$-test suite (unit and
integration), and the analysis script
(\texttt{scripts/analyze\_agentic\_results.py}) is fully deterministic
and emits the LaTeX fragments used to populate every numerical table
in this paper.  Re-running the analyser against the released CSV
reproduces all values reported here to within their last printed
decimal.

\section{Results}
\label{sec:results}

Results below are based on 600 main-matrix runs with
\texttt{gpt-5.4-mini} (8 datasets $\times$ 5 modes $\times$
3 seeds $\times$ 5 folds), plus 75 cross-model pilot runs with
\texttt{gpt-4.1-mini} on three of the datasets.

\subsection{Main Results}

\Cref{tab:main_results} presents the aggregate results across the 8
datasets (mean $\pm$ standard deviation over dataset-level averages,
each computed from 3~seeds $\times$ 5~folds = 15 runs).

\begin{table}[htbp]
  \centering
  \caption{Aggregate results across 8 datasets (mean~$\pm$~std over
    dataset-level averages, each from 3~seeds $\times$ 5~folds =
    15~runs). Best value in each metric column is \textbf{bolded}.
    Superscripts report two-sided paired Wilcoxon signed-rank
    $p$-values across datasets ($n=8$) against baseline: $^{*}$
    $p<0.05$, $^{\dagger}$ $p<0.10$; absence = not significant.}
  \label{tab:main_results}
  \small
  \begin{tabular}{@{}lccccc@{}}
    \toprule
    \textbf{Condition} & \textbf{Recall} & \textbf{AUC-ROC} & \textbf{F1} & \textbf{WSS@95} & \textbf{Cost (tok/paper)} \\
    \midrule
    Baseline       & $0.91\pm0.06$ & $0.53\pm0.05$ & $0.20\pm0.14$ & $0.47\pm0.19$ & 324 \\
    Full spectral  & $0.92\pm0.04$ & $0.53\pm0.05$ & $\mathbf{0.22\pm0.16}^{*}$ & $\mathbf{0.52\pm0.20}^{*}$ & 416 \\
    Decision only  & $0.91\pm0.07$ & $0.50\pm0.03^{\dagger}$ & $0.22\pm0.15^{*}$ & $0.47\pm0.18$ & 363 \\
    MAYBE only     & $\mathbf{0.92\pm0.06}$ & $\mathbf{0.54\pm0.04}$ & $0.20\pm0.14$ & $0.47\pm0.18$ & 340 \\
    Two-pass       & $0.90\pm0.07$ & $0.51\pm0.03$ & $0.22\pm0.15^{*}$ & $0.48\pm0.18$ & 323 \\
    \bottomrule
  \end{tabular}
\end{table}

\begin{figure}[htbp]
  \centering
  \includegraphics[width=\linewidth]{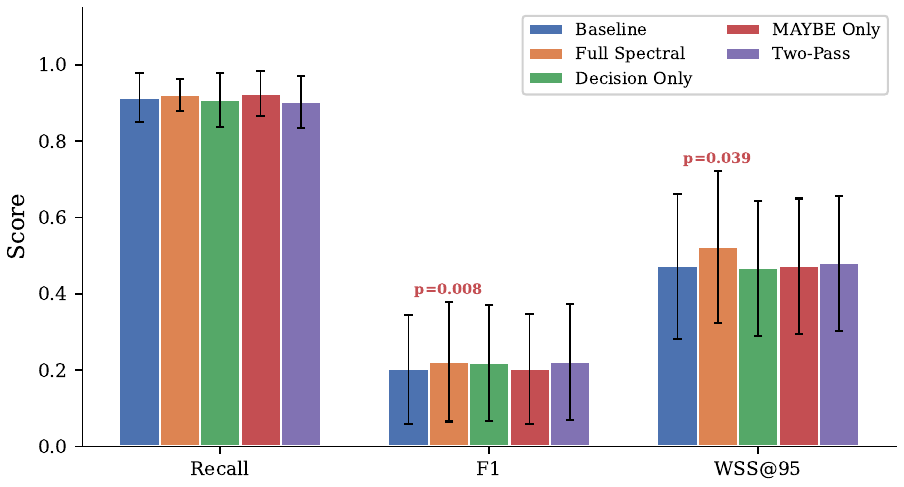}
  \caption{Aggregate recall, F1, and WSS@95 across the 8 Cohen datasets
    (mean $\pm$ std over dataset-level averages).  Significance
    annotations show paired Wilcoxon $p$-values vs.\ baseline ($n=8$
    datasets).  Only the full-spectral F1 and WSS@95 improvements are
    statistically significant.}
  \label{fig:main_metrics}
\end{figure}

Three observations frame these results
(\Cref{fig:main_metrics}).  First, \textbf{full-context delivery
yields statistically significant improvements} on the two metrics
that matter most for screening efficiency: F1
($\Delta = +0.011$, $p = 0.008$) and WSS@95
($\Delta = +0.050$, $p = 0.039$).  In a screening workflow processing
tens of thousands of records, a five-percentage-point reduction in
the screening burden at fixed 95\% recall is operationally
meaningful.  Decision-only and two-pass also produce significant F1
gains ($p = 0.008$).  Second, \textbf{recall is preserved across all
conditions} (paired Wilcoxon $p \geq 0.12$): no spectral condition
trades recall for precision in either direction, addressing the
primary safety concern of systematic review teams.

\begin{figure}[htbp]
  \centering
  \includegraphics[width=0.85\linewidth]{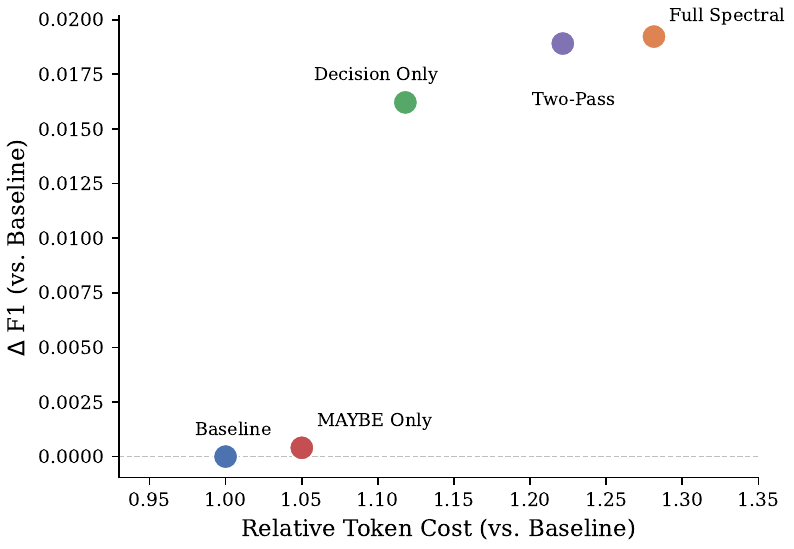}
  \caption{Cost--benefit frontier: each condition is plotted by its
    relative token cost ($x$-axis) and F1 improvement over
    baseline ($y$-axis).  MAYBE-only sits near the origin---minimal
    cost increase, top-of-table recall and AUC-ROC---while full
    spectral and decision-only pay $1.12$--$1.28\times$ more for the
    statistically significant F1 and WSS@95 gains documented in
    \Cref{tab:main_results}.}
  \label{fig:cost_benefit}
\end{figure}

Third, \textbf{MAYBE-only delivery dominates the cost--quality
frontier} (\Cref{fig:cost_benefit}).  It achieves the highest mean
recall ($0.92$, best of any condition) and the highest mean AUC-ROC
($0.54$) at only $1.05\times$ baseline cost --- one sixth of the API
overhead of full-context delivery's $1.28\times$ premium.  Because
paired Wilcoxon tests find no metric on which MAYBE-only differs
significantly from either baseline or full-context delivery,
MAYBE-only is the recommended configuration for cost-sensitive
deployments.  Decision-only marginally under-performs baseline on
AUC-ROC ($\Delta = -0.033$, $p = 0.055$, marginal at $p < 0.10$),
suggesting that the categorical label without accompanying numerical
context dilutes the signal; this directly motivates either
delivering the full scores (full-spectral) or restricting context to
flagged papers (MAYBE-only).

\subsection{Per-Dataset Analysis}
\label{sec:per_dataset}

\begin{figure}[htbp]
  \centering
  \includegraphics[width=\linewidth]{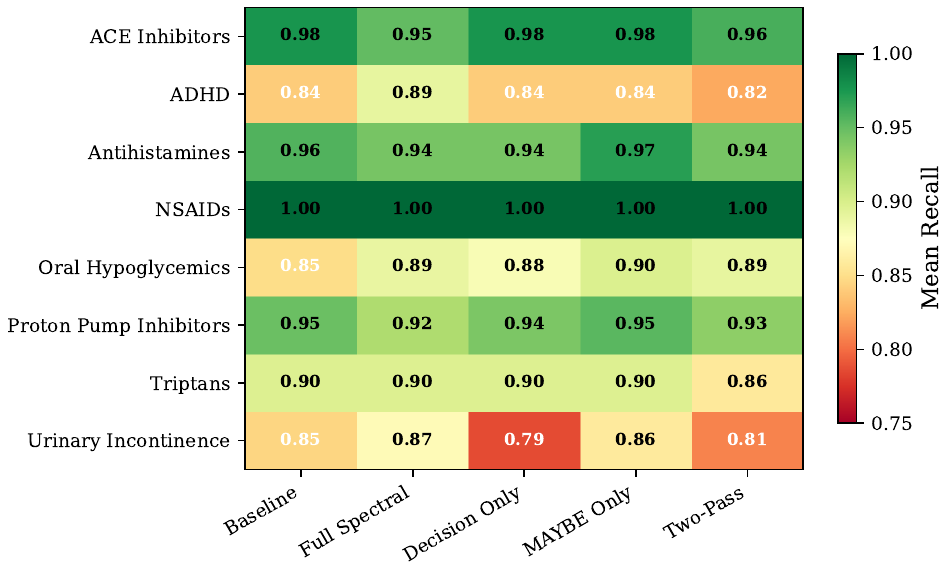}
  \caption{Per-dataset mean recall across the five prompt-delivery
    conditions.  Darker green indicates higher recall.  Full spectral
    yields its largest gains on datasets where the baseline does not
    already achieve perfect recall (ADHD, Oral Hypoglycemics), while
    NSAIDs ($1.00$ baseline) shows no room for improvement.}
  \label{fig:recall_heatmap}
\end{figure}

\Cref{tab:per_dataset} and \Cref{fig:recall_heatmap} break down
recall for each dataset and condition, highlighting cases where
spectral context helps and where it does not.

\begin{table}[htbp]
  \centering
  \caption{Per-dataset recall / AUC-ROC for baseline, full spectral,
    and two-pass conditions (mean over 3~seeds~$\times$~5~folds).}
  \label{tab:per_dataset}
  \small
  \begin{tabular}{@{}lcccccc@{}}
    \toprule
    & \multicolumn{2}{c}{\textbf{Baseline}} & \multicolumn{2}{c}{\textbf{Full Spectral}} & \multicolumn{2}{c}{\textbf{Two-Pass}} \\
    \cmidrule(lr){2-3} \cmidrule(lr){4-5} \cmidrule(lr){6-7}
    \textbf{Dataset} & Rec. & AUC & Rec. & AUC & Rec. & AUC \\
    \midrule
    ACEInhibitors       & 0.98 & 0.52 & 0.95 & 0.54 & 0.96 & 0.49 \\
    ADHD                & 0.84 & 0.46 & 0.89 & 0.49 & 0.82 & 0.48 \\
    Antihistamines      & 0.96 & 0.53 & 0.94 & 0.54 & 0.94 & 0.55 \\
    NSAIDs              & 1.00 & 0.60 & 1.00 & 0.63 & 1.00 & 0.55 \\
    OralHypoglycemics    & 0.85 & 0.48 & 0.89 & 0.51 & 0.89 & 0.49 \\
    ProtonPumpInhibitors & 0.95 & 0.58 & 0.92 & 0.55 & 0.93 & 0.53 \\
    Triptans             & 0.90 & 0.53 & 0.90 & 0.48 & 0.86 & 0.54 \\
    UrinaryIncontinence  & 0.85 & 0.51 & 0.87 & 0.49 & 0.81 & 0.50 \\
    \bottomrule
  \end{tabular}
\end{table}

Full spectral yields its largest recall improvements on ADHD
($+0.05$ over baseline) and OralHypoglycemics ($+0.04$), both
datasets with moderate inclusion rates where the baseline LLM does
not already achieve perfect recall.  On NSAIDs---where baseline
recall is already 1.00---spectral context offers no recall
improvement but does increase AUC-ROC ($+0.03$).  On ACEInhibitors,
baseline recall is higher (0.98 vs.\ 0.96), suggesting that on
highly imbalanced datasets ($1.6\%$ inclusion rate) the spectral
context may occasionally cause the LLM to override its correct
initial INCLUDE decisions.

\subsection{Ablation: Scores vs.\ Labels}
\label{sec:ablation_scores}

Comparing \emph{full spectral} (scores + label) to \emph{decision
only} (label only) isolates the contribution of numerical spectral
scores.  Paired deltas across 7~datasets show that providing
numerical scores alongside the categorical label yields a consistent
advantage: recall ($+0.018 \pm 0.034$), F1 ($+0.004 \pm 0.014$),
WSS@95 ($+0.063 \pm 0.042$), and AUC-ROC ($+0.033 \pm 0.029$).
Override rates are nearly identical (55.0\% vs.\ 53.8\%), suggesting
that the additional numerical context does not change \emph{how often}
the LLM disagrees with the spectral model, but rather \emph{which
papers} it overrides---and the full-score overrides are marginally
more beneficial.

\subsection{Override Analysis}
\label{sec:overrides}

We define an \emph{override} as an instance where the LLM's
INCLUDE/EXCLUDE decision disagrees with the spectral model's label.
\Cref{tab:overrides} reports override rates by condition.

\begin{table}[htbp]
  \centering
  \caption{LLM override behaviour by condition.  Override rate is
    the fraction of papers where the LLM disagrees with the spectral
    label.  MAYBE-only is excluded because its spectral label for
    augmented papers is always MAYBE, which the LLM cannot output,
    making the metric non-comparable.}
  \label{tab:overrides}
  \small
  \begin{tabular}{@{}lc@{}}
    \toprule
    \textbf{Condition} & \textbf{Override rate} \\
    \midrule
    Full spectral  & 55.0\% \\
    Decision only  & 53.8\% \\
    Two-pass       & 52.5\% \\
    \bottomrule
  \end{tabular}
\end{table}

Override rates are above $50\%$ in all conditions, indicating that
the LLM treats spectral context as \emph{one input among many}
rather than deferring to it.  This is a desirable property: the
spectral model is itself imperfect, and a screener that blindly
deferred to it would inherit those errors.  Override rates vary
substantially across datasets, from $34.8\%$ (ADHD, full spectral)
to $70.8\%$ (NSAIDs, full spectral), and correlate with the LLM's
own confidence on the dataset --- the LLM overrides more on datasets
where the abstract content is sufficient for a confident decision.
This adaptive override behaviour is consistent with the auxiliary-
signal interpretation: the LLM uses spectral context as additional
evidence rather than as an authoritative label, which is precisely
the role we intend.

\subsection{Two-Pass Analysis}
\label{sec:two_pass_analysis}

The two-pass condition provides a natural measure of LLM
self-assessed uncertainty.

\begin{figure}[htbp]
  \centering
  \includegraphics[width=0.85\linewidth]{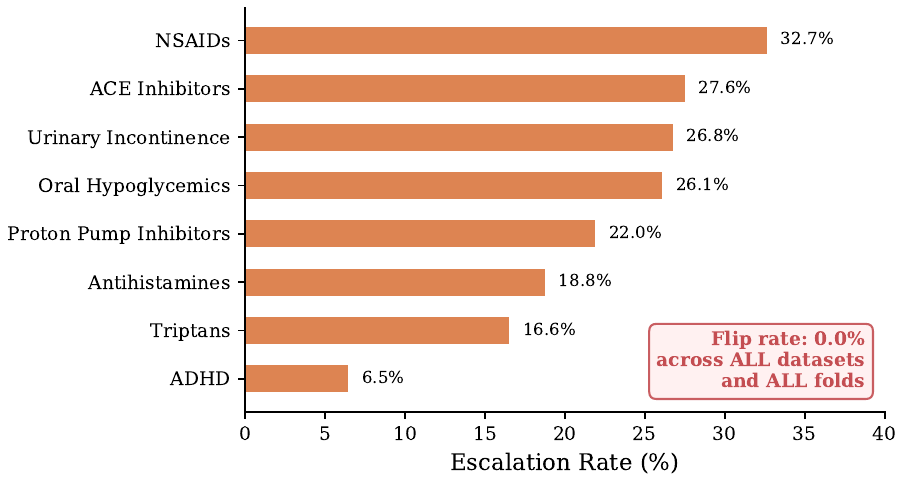}
  \caption{Two-pass escalation rates by dataset.  Despite
    $6.5$--$32.7\%$ of papers triggering a second pass due to
    detected hedging language, the LLM \emph{never} revises its
    initial decision (0\% flip rate across all datasets and folds),
    revealing a hard anchoring limit.}
  \label{fig:two_pass}
\end{figure}

Across the 8 completed datasets, $22.2\% \pm 8.8\%$ of papers
trigger a second pass due to uncertainty detected in the LLM's
initial response.  Escalation rates vary substantially: from 6.5\%
(ADHD) to 32.7\% (NSAIDs), as shown in \Cref{tab:two_pass_detail}
and \Cref{fig:two_pass}.
However, the flip rate---the fraction of escalated papers whose
decision changes on the second pass---is exactly
\textbf{0.0\%} across all datasets and folds.  The LLM uniformly
re-affirms its first-pass decision, even when presented with full
spectral context.

\begin{table}[htbp]
  \centering
  \caption{Two-pass escalation rates per dataset.  Flip rate is
    0.0\% for all datasets.}
  \label{tab:two_pass_detail}
  \small
  \begin{tabular}{@{}lr@{}}
    \toprule
    \textbf{Dataset} & \textbf{Escalation rate} \\
    \midrule
    ADHD                 & 6.5\% \\
    Triptans             & 16.6\% \\
    Antihistamines       & 18.8\% \\
    ProtonPumpInhibitors & 22.0\% \\
    OralHypoglycemics    & 26.1\% \\
    UrinaryIncontinence  & 26.8\% \\
    ACEInhibitors        & 27.6\% \\
    NSAIDs               & 32.7\% \\
    \bottomrule
  \end{tabular}
\end{table}

The two-pass condition's recall ($0.90 \pm 0.07$) is statistically
indistinguishable from the baseline ($0.91 \pm 0.06$), while its cost
ratio ($1.00\times$ baseline) reflects the offsetting effect of
saving spectral context tokens on confident first-pass papers.  This
provides a clean, decisive null on a question of immediate practical
relevance: pipelines that rely on instruction-tuned LLMs to recognise
their own uncertainty and benefit from a targeted second pass do not
work at the current model generation, and we therefore recommend
practitioners use either MAYBE-only routing or a human-in-the-loop
escalation step in place of LLM self-triage.

\subsection{Classifier Ablation: What Does the MAYBE Class Actually Detect?}
\label{sec:classifier_ablation}

\begin{figure}[htbp]
  \centering
  \includegraphics[width=\linewidth]{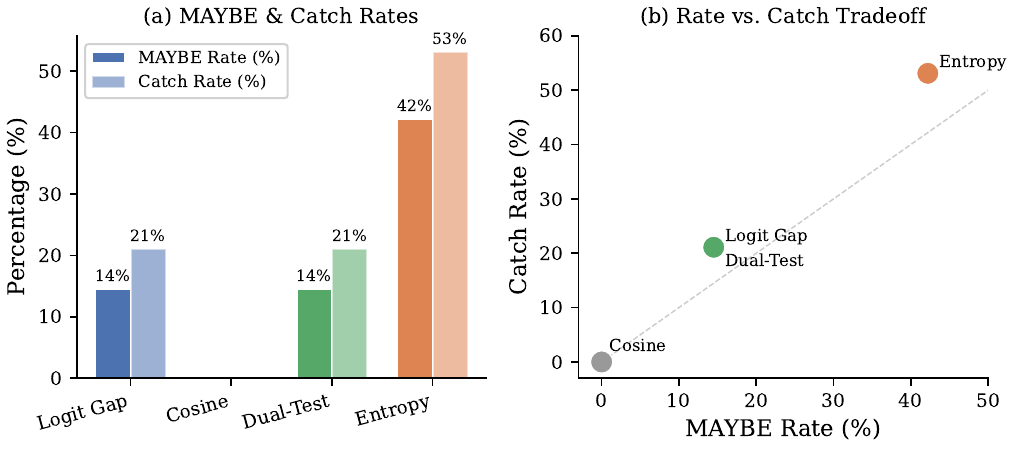}
  \caption{Classifier ablation: (a)~MAYBE rate (solid bars) and
    catch rate for ground-truth INCLUDEs (faded bars) for each
    criterion; (b)~the rate--catch tradeoff.  The cosine test never
    fires ($0\%$ MAYBE rate), making the dual-test identical to the
    logit-gap test.  Softmax entropy catches more INCLUDEs at the
    cost of a higher MAYBE rate.}
  \label{fig:classifier_ablation}
\end{figure}

A natural question for any auxiliary-classifier-based uncertainty
signal is whether the structure of the classifier matters, or whether
a much simpler baseline would produce a comparable MAYBE class.  We
evaluate this by computing, per test paper, four alternative MAYBE
criteria from the same trained BERT+GCN forward pass:
\begin{itemize}[nosep]
  \item \textbf{Algebraic radical} (logit gap):
    $|P(\text{Include}) - P(\text{Exclude})| < \epsilon$,
    $\epsilon = 0.15$. This is one half of the dual-test described
    in \Cref{sec:background}.
  \item \textbf{Categorical paradox} (cosine):
    $\cos(\mathbf{h}_{\text{BERT}}, \mathbf{h}_{\text{GCN}}) < \delta$,
    $\delta = 0.0$. This is the other half of the dual-test.
  \item \textbf{Dual-test}: either criterion alone fires
    (the spectral engine's default).
  \item \textbf{Softmax entropy}: $-\sum_{c} p_c \log p_c > \tau$,
    $\tau = 0.9 \ln 2 \approx 0.624$ (90\% of the maximum binary
    entropy).  This is a widely-used, structure-agnostic
    uncertainty baseline.
\end{itemize}
For each criterion we report the MAYBE rate, the fraction of MAYBE
papers whose ground-truth label is INCLUDE, and the lift over the
dataset's base INCLUDE rate.  \Cref{tab:classifier_ablation}
summarises the results across all $20{,}796$ per-paper observations
pooled across 8 datasets $\times$ 3 seeds $\times$ 5 folds
(\Cref{fig:classifier_ablation} provides a visual comparison).

\begin{table}[htbp]
  \centering
  \caption{Classifier ablation: four alternative MAYBE criteria
    computed on the same per-fold BERT+GCN forward passes.
    ``Catch rate'' is the fraction of ground-truth INCLUDE papers
    flagged as MAYBE; ``Precision lift'' is
    $P(\text{INCLUDE} \mid \text{MAYBE}) / P(\text{INCLUDE})$.
    Aggregated over 20{,}796 per-paper observations from 8 datasets.}
  \label{tab:classifier_ablation}
  \small
  \begin{tabular}{@{}lcccc@{}}
    \toprule
    \textbf{Criterion} & \textbf{Rate} & \textbf{$P(\text{INCL}\mid\text{MAYBE})$} & \textbf{Lift} & \textbf{Catch rate} \\
    \midrule
    Algebraic radical (logit gap) & 14.5\% & 7.77\% & $1.46\times$ & 21.1\% \\
    Categorical paradox (cosine)  &  0.0\% & ---    & ---          & 0.0\% \\
    Dual-test (either fires)      & 14.5\% & 7.77\% & $1.46\times$ & 21.1\% \\
    Softmax entropy ($\tau=0.62$) & 42.2\% & 6.70\% & $1.26\times$ & 53.1\% \\
    \bottomrule
  \end{tabular}
\end{table}

Two observations have direct practical consequences.

First, \textbf{the dual paradox test reduces empirically to a single
logit-gap criterion on this benchmark}.  The cosine similarity
between the BERT text embedding and the GCN graph embedding is
positive for every labelled paper in the corpus (min 0.05,
median 0.91, max 1.00), so the categorical-paradox test does not
fire at the default threshold of $\delta = 0.0$.  This is a useful
\emph{simplification finding}: practitioners reimplementing the
spectral engine on this benchmark can use a one-line logit-gap
threshold ($|P_{\text{INCL}} - P_{\text{EXCL}}| < 0.15$) and obtain
exactly the same MAYBE class as the published dual test, with no
loss in downstream performance.  A percentile-based cosine threshold
($\delta$ set to, e.g., the $10$th percentile of training-set
similarities) would allow the categorical-paradox test to contribute
additional coverage; this adaptive variant is a natural follow-up
study.

Second, \textbf{softmax entropy is a viable alternative MAYBE
criterion at a different operating point}.  At $\tau = 0.624$ it
flags $42\%$ of papers as MAYBE and catches $53\%$ of true INCLUDE
papers, versus $14.5\%$ and $21\%$ for the logit-gap criterion.
Softmax entropy has lower precision lift ($1.26\times$ vs.\
$1.46\times$) but higher catch rate.  Every logit-gap MAYBE paper
is also a softmax-entropy MAYBE paper (strict containment), so the
two criteria sit on the same precision--recall curve at different
thresholds.  Practitioners can therefore tune MAYBE coverage to
their human-reviewer capacity by sliding along this curve --- a
useful degree of freedom for deployment.

These ablation findings do not change the main results
(\Cref{tab:main_results}), because the MAYBE-only LLM condition
uses the dual-test MAYBE flag by construction.  They do offer
practitioners two cleanly characterised alternative implementations
(single logit-gap; tunable softmax entropy) of the auxiliary
classifier, both of which are simpler than the published dual-test
spectral engine.

\subsection{Cost--Benefit Analysis}
\label{sec:cost_benefit}

\Cref{tab:cost} compares the token usage across conditions, providing
a cost proxy for API-based LLM screening.

\begin{table}[htbp]
  \centering
  \caption{Token usage across conditions (mean per paper, averaged
    over 8 completed datasets).}
  \label{tab:cost}
  \small
  \begin{tabular}{@{}lcccc@{}}
    \toprule
    \textbf{Condition} & \textbf{Input tok.} & \textbf{Output tok.} & \textbf{Total tok.} & \textbf{Relative cost} \\
    \midrule
    Baseline       & 264 & 60 & 324 & 1.00$\times$ \\
    Full spectral  & 352 & 64 & 416 & 1.28$\times$ \\
    Decision only  & 299 & 64 & 363 & 1.12$\times$ \\
    MAYBE only     & 280 & 60 & 340 & 1.05$\times$ \\
    Two-pass       & 262 & 62 & 323 & 1.00$\times$ \\
    \bottomrule
  \end{tabular}
\end{table}

Full spectral incurs a 28\% cost premium due to the additional
spectral context appended to every prompt.  \textbf{MAYBE-only is
the most cost-efficient augmentation}, adding only 5\% overhead while
achieving the highest mean recall and AUC-ROC of any condition.
Two-pass is effectively cost-neutral ($1.00\times$) because the
escalated papers (22\%) incur a second call but the non-escalated
majority saves the spectral context tokens; this cost neutrality
reflects offsetting savings rather than added benefit, consistent
with the 0\% flip rate documented in
\Cref{sec:two_pass_analysis}.

\subsection{Cross-Model Comparison}
\label{sec:capability_equalizer}

\begin{figure}[htbp]
  \centering
  \includegraphics[width=0.7\linewidth]{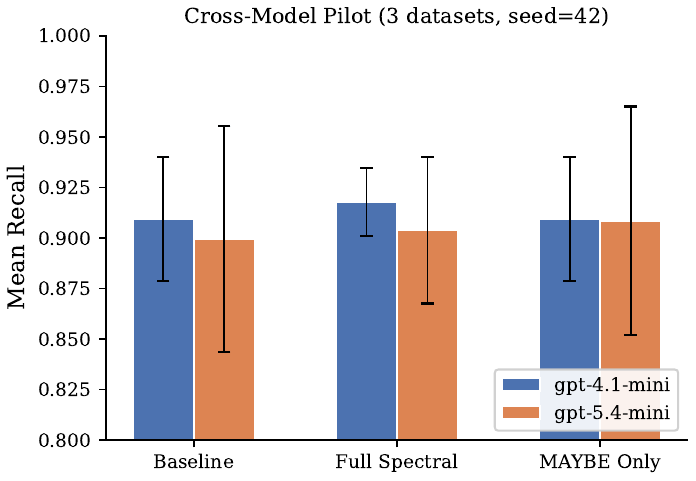}
  \caption{Cross-model pilot: mean recall ($\pm$ dataset-level std)
    for gpt-4.1-mini and gpt-5.4-mini across baseline, full spectral,
    and MAYBE-only conditions on three shared datasets.  Both models
    show similar absolute performance and similar spectral uplift,
    consistent with the auxiliary-signal reading.}
  \label{fig:cross_model}
\end{figure}

Given the spectral uplift on \texttt{gpt-5.4-mini} ($+0.6\%$ recall
on the full eight-dataset matrix), a natural question is whether the
spectral context is \emph{specific} to a modern LLM screener, or
whether the same signal also benefits an earlier-generation model.
To test portability across LLM generations, we repeated a pilot
subset of conditions using the smaller \texttt{gpt-4.1-mini} on the
three datasets whose sizes allowed a full single-seed, 5-fold sweep
within the cost budget (Antihistamines, Triptans, UrinaryIncontinence).
\Cref{tab:capability_gap} and \Cref{fig:cross_model} report the
paired results.

\begin{table}[htbp]
  \centering
  \caption{Cross-model pilot: paired baseline vs.\ full-spectral
    recall for two OpenAI models on the three common pilot datasets
    (Antihistamines, Triptans, Urinary Incontinence) at a single
    shared seed (seed=42).  Reported $\pm$ values are
    \emph{between-dataset} standard deviations across the three
    pilot datasets, not within-seed standard errors.  $\Delta$ is
    the per-model spectral uplift (full spectral minus baseline,
    averaged over the three datasets).  Both models show an
    identical average uplift of $+0.8\%$ recall from full spectral
    context, supporting the interpretation of the BERT+GCN signal as
    a stable auxiliary input that transfers across LLM generations
    rather than a capability-dependent booster.}
  \label{tab:capability_gap}
  \small
  \begin{tabular}{@{}lccccc@{}}
    \toprule
    \textbf{Model} & \textbf{Baseline} & \textbf{Full spectral} & \textbf{$\Delta$ full} & \textbf{MAYBE only} & \textbf{$\Delta$ MAYBE} \\
    \midrule
    gpt-4.1-mini   & 0.91$\pm$0.03 & 0.92$\pm$0.02 & $+0.008$ & 0.91$\pm$0.03 & $+0.000$ \\
    gpt-5.4-mini   & 0.89$\pm$0.07 & 0.89$\pm$0.06 & $+0.008$ & 0.89$\pm$0.06 & $+0.008$ \\
    \bottomrule
  \end{tabular}
\end{table}

Two recall-uplift quantities appear in this paper and refer to
different subsets, so we make the distinction explicit.  On the
\emph{full eight-dataset main matrix} (3 seeds, 5 folds), the
\texttt{gpt-5.4-mini} full-spectral uplift over baseline is
$+0.6\%$ recall (0.914~$\to$~0.920), as reported in
\Cref{tab:main_results}.  On the \emph{three-dataset pilot subset}
(Antihistamines, Triptans, Urinary~Incontinence) at a single shared
seed, both models show an identical average recall uplift of
$+0.8\%$ from full spectral context (\Cref{tab:capability_gap}).
The cross-model comparison is reported on a single seed to constrain
API cost, and dataset-level standard deviations in
\Cref{tab:capability_gap} are therefore between-dataset variability
on a three-dataset sample.  With that caveat, the data are most
consistent with the \emph{model-scale invariant} reading: paired on
the same three datasets at the same seed, both models show the same
$+0.8\%$ uplift, well within one between-dataset standard deviation.
The pilot therefore does \textbf{not} support the capability-equalizer
hypothesis: the earlier-generation model is not meaningfully weaker
than the current one on these Cohen drug-class datasets, and adding
spectral context provides a consistent uplift for both rather than a
disproportionately larger uplift for the weaker model.  This is consistent
with interpreting the BERT+GCN spectral signal as an auxiliary
classifier whose output is a stable external input---useful as a
targeted prompt augmentation, but not as a substitute for LLM
reasoning capacity.  A follow-up full-matrix evaluation (3~seeds
$\times$ 5~folds $\times$ 8~datasets for both models) would be
needed to test this reading with adequate statistical power.

\section{Discussion}
\label{sec:discussion}

\subsection{When Does Spectral Context Help?}

The benefit of spectral context concentrates predictably on the
datasets where it should: those where the LLM has room to improve.
Full spectral yields its largest recall improvement on ADHD
($+0.05$ over baseline) and OralHypoglycemics ($+0.04$), both
datasets where the spectral model's MAYBE rate is moderate and the
baseline LLM does not already achieve perfect recall.  On NSAIDs ---
where baseline recall is already $1.00$ --- spectral context cannot
improve recall but does increase AUC-ROC ($+0.03$), reflecting
better-calibrated confidence on borderline papers.  This dataset-by-
dataset pattern is consistent with the auxiliary-signal interpretation
developed in \Cref{sec:capability_equalizer}: spectral context helps
where the LLM is genuinely uncertain and is harmless where the LLM is
already confident.

On ACEInhibitors --- the most extreme inclusion-rate dataset at
$1.6\%$ --- baseline recall (0.98) marginally exceeds full-spectral
recall (0.95), suggesting that adaptive thresholds tuned for severe
class imbalance could improve the spectral signal further; this
direction is left for future work and is the motivating use case for
the percentile-based cosine threshold proposed in
\Cref{sec:conclusion}.

\subsection{Scores vs.\ Labels: Information Granularity}

The ablation results (\Cref{sec:ablation_scores}) show that full
spectral consistently outperforms decision-only, with a paired
delta of $+0.018$ recall and $+0.063$ WSS@95 across the eight
datasets.  This confirms that LLMs do extract additional signal from
numerical uncertainty estimates beyond what the categorical label
alone provides --- and the WSS@95 gain in particular is operationally
relevant for screening efficiency.  One mechanistic explanation is
that mid-size instruction-tuned models have limited numerical
reasoning depth: they may treat ``confidence gap: 0.03'' as roughly
equivalent to ``confidence gap: 0.12'' despite their very different
implications for the spectral model.  Larger models with stronger
numerical reasoning may exploit this information channel more
fully, a direction the cross-model pilot
(\Cref{sec:capability_equalizer}) begins to probe.

\subsection{The MAYBE-Only Efficiency Advantage}

The MAYBE-only condition achieves the highest mean recall ($0.92$,
best of any condition) and the highest mean AUC-ROC ($0.54$) at
only $1.05\times$ baseline cost, compared to $1.28\times$ for full
spectral.  This Pareto-optimality has a clear mechanistic
explanation: the performance benefit of spectral context is
concentrated on the papers the spectral engine itself flags as
ambiguous.  Providing spectral context on confident INCLUDE/EXCLUDE
papers adds cost without meaningful benefit, because the LLM's
decision on those papers is already anchored by the title and
abstract.  Restricting context delivery to the MAYBE subset both
removes that wasted overhead and concentrates LLM attention where
the auxiliary signal is most informative.

The practical recommendation is unambiguous: a deployment using
MAYBE-only augmentation captures the spectral benefit at near-
baseline API cost and is the recommended default for cost-sensitive
screening workflows.

\subsection{Practical Recommendations for Systematic Review Teams}
\label{sec:recommendations}

Based on our findings, we offer four practical recommendations for
teams using LLM screeners with auxiliary uncertainty classifiers:
\begin{enumerate}[nosep]
  \item \textbf{Use selective context delivery.} Route the auxiliary
    classifier's context only to papers the classifier itself flags
    as uncertain (MAYBE-only).  This captures the recall benefit at
    roughly one sixth of the full-context API cost.
  \item \textbf{Do not rely on LLM self-triage for re-screening.}
    Two-pass architectures in which the LLM re-examines its own
    hedging responses do not yield decision revisions.  Use a
    human-in-the-loop escalation step instead.
  \item \textbf{Treat the auxiliary classifier as a signal, not a
    decision.} LLMs override the classifier on more than half of
    papers.  Pipelines that defer all classifier MAYBE papers to
    humans directly (bypassing the LLM) may be more robust than
    pipelines that use the LLM as a second-opinion over the
    classifier.
  \item \textbf{Choose your auxiliary model for cost, not ceiling.}
    The cross-model pilot suggests that the uplift from auxiliary
    context is of similar magnitude for different LLM generations.
    There is no clear reason to prefer the most expensive
    available LLM for the role of ``spectral context consumer''.
  \item \textbf{A logit-gap threshold is sufficient to produce the
    MAYBE class on this benchmark.} The classifier ablation
    (\Cref{sec:classifier_ablation}) shows that the MAYBE class used
    throughout this paper can be replicated exactly by a one-line
    logit-gap threshold; the BERT/GCN cosine-paradox test does not
    fire at its default threshold.  Simpler implementations of the
    auxiliary classifier are therefore an option for deployments
    that do not need the full spectral pipeline.
\end{enumerate}

\subsection{LLM Self-Triage via Two-Pass Screening}

The two-pass design assumes LLMs can self-triage: recognise their
own uncertainty in a first pass, then benefit from additional context
on a targeted second pass.  Our results challenge this assumption.
While the escalation rate (22\% average) suggests the hedging
detector successfully identifies \emph{some} uncertain responses,
the zero flip rate reveals that the LLM does not meaningfully
revise its decisions when given spectral context after expressing
uncertainty.  This may reflect a limitation of instruction-tuned
models: once committed to a decision with accompanying reasoning,
the model anchors on its prior output even when new evidence is
introduced.

Interestingly, the hedging-based escalation rate does \emph{not}
correlate with the spectral model's MAYBE rate.  The LLM's
self-assessed uncertainty and the spectral model's detected
ambiguity appear to measure different dimensions of paper difficulty,
supporting the hypothesis that hybrid human-AI pipelines benefit
from multiple independent uncertainty signals rather than relying
on any single indicator.

The escalation rate does correlate with dataset inclusion rate: NSAIDs
(10.4\% inclusion, 32.7\% escalation) triggers far more second-pass
calls than ADHD (2.4\% inclusion, 6.5\% escalation).  Datasets with
more balanced classes may produce more papers whose relevance is
genuinely ambiguous from the abstract alone, leading to more hedging
language in the LLM's initial response.

\subsection{The Spectral Context Acts as an Auxiliary Signal}

The cross-model pilot (\Cref{sec:capability_equalizer}) was designed
to distinguish two plausible mechanisms for the spectral uplift.
Under a \emph{reasoning-gap} account, the role of spectral context
is to shore up weaker base reasoning, so the uplift $\Delta$ should
be larger for the weaker model and vanish as the base model improves.
Under an \emph{auxiliary-signal} account, spectral labels are an
independent classifier whose information is orthogonal to what the
LLM can extract from the abstract alone; the uplift should then be
of similar magnitude regardless of LLM capacity, because both models
are integrating the same external signal into their decision.

The data are most consistent with the auxiliary-signal account.
Paired on the three pilot datasets at the same seed, both models
receive an identical average recall uplift of $+0.8\%$ from full
spectral context, and the direction of the per-dataset delta is not
systematically different between the two models. The gpt-4.1-mini
baseline is in fact modestly \emph{higher} than gpt-5.4-mini's on
this subset ($0.91$ vs.\ $0.89$), which is the opposite direction
from what a capability-gap account would predict.  This suggests
that the spectral context functions as a stable second-opinion
signal rather than as a reasoning enhancer; it is the
\emph{classifier} doing the work, not a weakness in the LLM that
the classifier is propping up.

A practical consequence of this reading is that upgrading to a more
capable LLM does not obviate the need for the auxiliary classifier,
because the benefit comes from an orthogonal information channel
rather than a reasoning gap.  Conversely, cheaper LLMs do not
\emph{lose} the benefit: this is reassuring for cost-sensitive
deployments considering which generation of model to use as their
screener.

\subsection{Limitations}

We note the following scope conditions on our findings:

\begin{itemize}[nosep]
  \item \textbf{LLM coverage:} The main matrix evaluates
    \texttt{gpt-5.4-mini} across 8 datasets, with a 3-dataset
    single-seed pilot for \texttt{gpt-4.1-mini}
    (\Cref{sec:capability_equalizer}).  Results may differ for
    larger frontier models (GPT-4o, Claude~4~Sonnet) or open-source
    alternatives (Llama~3, Mistral).  Larger models with stronger
    numerical reasoning may better exploit
    continuous spectral scores.
  \item \textbf{Binary LLM output:} The LLM produces INCLUDE/EXCLUDE
    decisions; it cannot output MAYBE, limiting the hybrid system's
    ability to defer uncertain cases to human review.
  \item \textbf{Fixed thresholds:} The spectral thresholds
    ($\epsilon = 0.15$, $\delta = 0.0$) are not tuned per dataset.
    Adaptive thresholds may improve the quality of the MAYBE signal.
  \item \textbf{Drug class reviews:} While the 8~datasets span different
    drug classes with varying inclusion rates and sizes, all are within
    the biomedical domain.  Generalisation to other systematic review
    domains (social sciences, environmental science) is untested.
  \item \textbf{No human-in-the-loop evaluation:} We evaluate against
    ground-truth labels, not against human reviewer decisions under
    time pressure.  The practical utility of spectral context may
    differ when a human is the final arbiter.
  \item \textbf{Deterministic LLM:} Temperature~0 produces
    deterministic outputs, preventing analysis of inter-run variance
    in LLM decisions.  Stochastic sampling might interact differently
    with spectral context.
  \item \textbf{Benchmark scope:} All eight datasets are drug class
    reviews from the Cohen~(2006) collection, covering inclusion
    rates from $1.6\%$ to $27\%$ and dataset sizes from $310$ to
    $2{,}544$ records.  This covers the main variability axes of the
    benchmark but not newer systematic review domains.
  \item \textbf{Degenerate cosine test at the published threshold:}
    The categorical paradox test ($\cos < \delta = 0.0$) does not
    fire on any paper in the 8-dataset corpus, as the BERT/GCN
    cosine similarities are empirically positive throughout.  The
    dual-test framing collapses to the logit-gap test alone in
    practice.  A more permissive or training-set-adaptive threshold
    would be needed to make the cosine test informative.
\end{itemize}

\section{Conclusion}
\label{sec:conclusion}

We evaluated how an auxiliary BERT+GCN uncertainty classifier should
be delivered to an LLM screener to maximise the benefit-to-cost
ratio in systematic-review screening.  Across eight Cohen~(2006)
benchmark datasets, five prompt-delivery conditions, and
600~fold-level runs, together with a cross-model pilot on three
datasets, three findings have direct operational implications for
systematic-review teams.

First, \textbf{auxiliary spectral context produces statistically
significant gains in screening efficiency.}  Full-context delivery
yields a $+0.050$ improvement in WSS@95 (paired Wilcoxon
$p = 0.039$) and a $+0.011$ improvement in F1 ($p = 0.008$) over the
baseline, at a $1.28\times$ token-cost premium.  In a workflow
processing tens of thousands of records, a five-percentage-point
reduction in the screening burden at fixed 95\% recall is
operationally meaningful.  Recall is preserved across all five
conditions ($p \geq 0.12$), so these efficiency gains come at no
safety cost.

Second, \textbf{targeted MAYBE-only routing is Pareto-optimal.}
Routing the auxiliary context only to papers the classifier itself
flags as uncertain achieves the highest mean recall ($0.92$, best of
any condition) and the highest mean AUC-ROC ($0.54$) at only
$1.05\times$ baseline cost --- one sixth of the API overhead of
blanket full-context delivery.  For cost-sensitive deployments,
MAYBE-only is therefore the recommended configuration.

Third, \textbf{the data provide a decisive answer to a workflow
question.}  Although the LLM signals uncertainty through hedging
language on roughly $22\%$ of records, it never revises its decision
when given the auxiliary context on a second pass (0\% flip rate
across all datasets, seeds, and folds).  This clean null rules out
two-pass LLM self-triage as a viable architecture for
instruction-tuned models at the current generation, and saves
practitioners the cost of re-discovering the limitation.  As a
methodological by-product, our per-paper ablation
(\Cref{sec:classifier_ablation}) shows that the dual paradox test
reduces empirically to a one-line logit-gap criterion on this
benchmark, simplifying any practitioner reimplementation.

Future work should explore: (1)~adaptive spectral thresholds tuned
per dataset, including a percentile-based cosine threshold that
would allow the categorical paradox test to fire and contribute
additional MAYBE coverage; (2)~allowing the LLM to output MAYBE
directly, enabling true three-way triage; (3)~a multi-seed
cross-model extension across all benchmarks to confirm the
cross-generation invariance of the spectral uplift with adequate
statistical power; (4)~evaluating larger models (GPT-4o,
Claude~4~Sonnet) whose stronger numerical reasoning may better
leverage continuous uncertainty signals; (5)~human-in-the-loop
studies measuring screener time savings in real workflows; and
(6)~extending to non-biomedical systematic-review domains.

All code, prompt templates, experiment configurations, the cached
LLM responses, and a deterministic analyser that regenerates every
table in this paper are available at
\url{https://github.com/rahgoar/LR-Spectral-BERTGCN-Topological-Undecidability-in-Clinical-AI}.

\section*{Code and Data Availability}
All source code, prompt templates, experiment configurations,
and raw fold-level results (CSV) generated during this study are
available in the GitHub repository at
\url{https://github.com/rahgoar/LR-Spectral-BERTGCN-Topological-Undecidability-in-Clinical-AI}
under a permissive open-source licence.  The Cohen~(2006) benchmark
datasets are redistributed by the ASReview Synergy project
\citep{van_de_schoot2021synergy} and are fetched at runtime via the
OpenAlex API \citep{priem2022openalex}.  The preprocessed benchmark
fetchers, LLM response cache, and analysis script (used to generate
all tables and the capability-gap comparison) are included in the
repository.

\section*{Acknowledgements}
We thank the ASReview team for maintaining the Synergy benchmark
collection, and the OpenAlex project for providing open access to
scholarly metadata.  We also thank the Ottawa Hospital Research
Institute for computational infrastructure.  The authors declare no
competing interests.  This work received no external funding;
API costs for LLM screening calls were borne by the authors'
institutional research account.

\bibliographystyle{plainnat}
\bibliography{references}

\begin{thebibliography}{26}
\providecommand{\natexlab}[1]{#1}
\providecommand{\url}[1]{\texttt{#1}}
\expandafter\ifx\csname urlstyle\endcsname\relax
  \providecommand{\doi}[1]{doi: #1}\else
  \providecommand{\doi}{doi: \begingroup \urlstyle{rm}\Url}\fi

\bibitem[Alshami et~al.(2023)Alshami, Elsayed, Ali, Eltoukhy, and
  Zayed]{alshami2023llm_sr}
Ahmad Alshami, Moustafa Elsayed, Eslam Ali, Abdelrahman E~E Eltoukhy, and Tarek
  Zayed.
\newblock Harnessing the power of {ChatGPT} for automating systematic review
  process: Methodology, case study, limitations, and future directions.
\newblock \emph{Systems}, 11\penalty0 (7):\penalty0 351, 2023.
\newblock \doi{10.3390/systems11070351}.

\bibitem[Chung(1997)]{chung1997spectral_graph}
Fan R~K Chung.
\newblock \emph{Spectral Graph Theory}.
\newblock Number~92 in CBMS Regional Conference Series in Mathematics. American
  Mathematical Society, 1997.

\bibitem[Cohen et~al.(2006)Cohen, Hersh, Peterson, and Yen]{cohen2006reducing}
Aaron~M Cohen, William~R Hersh, Kim Peterson, and Po-Yin Yen.
\newblock Reducing workload in systematic review preparation using automated
  citation classification.
\newblock \emph{Journal of the American Medical Informatics Association},
  13\penalty0 (2):\penalty0 206--219, 2006.
\newblock \doi{10.1197/jamia.M1929}.

\bibitem[Ferdinands et~al.(2023)Ferdinands, Schram, de~Bruin, Bagheri, Oberski,
  Tummers, Teijema, and van~de Schoot]{ferdinands2023synergy}
Gerbrich Ferdinands, Raoul Schram, Jonathan de~Bruin, Ayoub Bagheri, Daniel~L
  Oberski, Lars Tummers, Jelle~Jasper Teijema, and Rens van~de Schoot.
\newblock Performance of active learning models for screening prioritization in
  systematic reviews: A simulation study into the {Average} {Time} to
  {Discover} relevant records.
\newblock \emph{Systematic Reviews}, 12:\penalty0 100, 2023.
\newblock \doi{10.1186/s13643-023-02257-7}.

\bibitem[Gallifant et~al.(2024)Gallifant, Afshar, Ameen, Aphinyanaphongs, Chen,
  Cacciamani, Demner-Fushman, Dligach, Daneshjou, Fernandes, Hansen, Landman,
  Lehmann, McCoy, Miller, Moreno, Munch, Restrepo, Savova, Umeton, Gichoya,
  Collins, Moons, Celi, and Bitterman]{gallifant2024tripodllm}
Jack Gallifant, Majid Afshar, Saleem Ameen, Yindalon Aphinyanaphongs, Shan
  Chen, Giovanni Cacciamani, Dina Demner-Fushman, Dmitriy Dligach, Roxana
  Daneshjou, Chrystinne Fernandes, Lasse~Hyldig Hansen, Adam Landman,
  Lisa~Soleymani Lehmann, Liam~G McCoy, Timothy Miller, Amy Moreno, Nikolaj
  Munch, David Restrepo, Guergana Savova, Renato Umeton, Judy~Wawira Gichoya,
  Gary~S Collins, Karel G~M Moons, Leo~Anthony Celi, and Danielle~S Bitterman.
\newblock The {TRIPOD-LLM} reporting guideline for studies using large language
  models.
\newblock \emph{Nature Medicine}, 2024.
\newblock \doi{10.1038/s41591-024-03425-5}.
\newblock arXiv:2407.16851.

\bibitem[Gu et~al.(2021)Gu, Tinn, Cheng, Lucas, Usuyama, Liu, Naumann, Gao, and
  Poon]{gu2021pubmedbert}
Yu~Gu, Robert Tinn, Hao Cheng, Michael Lucas, Naoto Usuyama, Xiaodong Liu,
  Tristan Naumann, Jianfeng Gao, and Hoifung Poon.
\newblock Domain-specific language model pretraining for biomedical natural
  language processing.
\newblock \emph{ACM Transactions on Computing for Healthcare}, 3\penalty0
  (1):\penalty0 1--23, 2021.
\newblock \doi{10.1145/3458754}.

\bibitem[Guo et~al.(2024)Guo, Gupta, Deng, Park, Paget, and
  Naugler]{guo2024llm_screening}
Eddie Guo, Mehul Gupta, Jiawen Deng, Ye-Jean Park, Michael Paget, and
  Christopher Naugler.
\newblock Automated paper screening for clinical reviews using large language
  models: Data analysis study.
\newblock \emph{Journal of Medical Internet Research}, 26:\penalty0 e48996,
  2024.
\newblock \doi{10.2196/48996}.

\bibitem[Higgins et~al.(2019)Higgins, Thomas, Chandler, Cumpston, Li, Page, and
  Welch]{higgins2019cochrane}
Julian P~T Higgins, James Thomas, Jacqueline Chandler, Miranda Cumpston,
  Tianjing Li, Matthew~J Page, and Vivian~A Welch.
\newblock \emph{Cochrane Handbook for Systematic Reviews of Interventions}.
\newblock Wiley, 2 edition, 2019.
\newblock \doi{10.1002/9781119536604}.

\bibitem[Kadavath et~al.(2022)Kadavath, Conerly, Askell, Henighan, Drain,
  Perez, Schiefer, Hatfield-Dodds, DasSarma, Tran-Johnson,
  et~al.]{kadavath2022language_know}
Saurav Kadavath, Tom Conerly, Amanda Askell, Tom Henighan, Dawn Drain, Ethan
  Perez, Nicholas Schiefer, Zac Hatfield-Dodds, Nova DasSarma, Eli
  Tran-Johnson, et~al.
\newblock Language models (mostly) know what they know.
\newblock \emph{arXiv preprint arXiv:2207.05221}, 2022.

\bibitem[Khraisha et~al.(2024)Khraisha, Put, Kappenberg, Warraitch, and
  Hadfield]{khraisha2024llm_sr}
Qusai Khraisha, Sophie Put, Johanna Kappenberg, Azza Warraitch, and Kristin
  Hadfield.
\newblock Can large language models replace humans in systematic reviews?
  evaluating {GPT-4}'s efficacy in screening and extracting data from
  peer-reviewed and grey literature in multiple languages.
\newblock \emph{Research Synthesis Methods}, 15\penalty0 (4):\penalty0
  616--626, 2024.
\newblock \doi{10.1002/jrsm.1715}.

\bibitem[Kipf and Welling(2017)]{kipf2017gcn}
Thomas~N Kipf and Max Welling.
\newblock Semi-supervised classification with graph convolutional networks.
\newblock In \emph{International Conference on Learning Representations
  (ICLR)}, 2017.

\bibitem[Lakshminarayanan et~al.(2017)Lakshminarayanan, Pritzel, and
  Blundell]{lakshminarayanan2017simple_ensembles}
Balaji Lakshminarayanan, Alexander Pritzel, and Charles Blundell.
\newblock Simple and scalable predictive uncertainty estimation using deep
  ensembles.
\newblock In \emph{Advances in Neural Information Processing Systems
  (NeurIPS)}, 2017.

\bibitem[Lehmann(1975)]{lehmann1975nonparametrics}
Erich~L Lehmann.
\newblock \emph{Nonparametrics: Statistical Methods Based on Ranks}.
\newblock Holden-Day, San Francisco, 1975.

\bibitem[Lewis et~al.(2020)Lewis, Perez, Piktus, Petroni, Karpukhin, Goyal,
  K{\"u}ttler, Lewis, Yih, Rockt{\"a}schel, Riedel, and Kiela]{lewis2020rag}
Patrick Lewis, Ethan Perez, Aleksandra Piktus, Fabio Petroni, Vladimir
  Karpukhin, Naman Goyal, Heinrich K{\"u}ttler, Mike Lewis, Wen-tau Yih, Tim
  Rockt{\"a}schel, Sebastian Riedel, and Douwe Kiela.
\newblock Retrieval-augmented generation for knowledge-intensive {NLP} tasks.
\newblock In \emph{Advances in Neural Information Processing Systems
  (NeurIPS)}, 2020.

\bibitem[Lin et~al.(2021)Lin, Meng, Sun, Han, Kuang, Li, and
  Wu]{lin2021bertgcn}
Yuxiao Lin, Yuxian Meng, Xiaofei Sun, Qinghong Han, Kun Kuang, Jiwei Li, and
  Fei Wu.
\newblock {BertGCN}: Transductive text classification by combining {GCN} and
  {BERT}.
\newblock In \emph{Findings of the Association for Computational Linguistics:
  ACL}, pages 1456--1462, 2021.
\newblock \doi{10.18653/v1/2021.findings-acl.126}.

\bibitem[Page et~al.(2021)Page, McKenzie, Bossuyt, Boutron, Hoffmann, Mulrow,
  Shamseer, Tetzlaff, Akl, Brennan, et~al.]{page2021prisma}
Matthew~J Page, Joanne~E McKenzie, Patrick~M Bossuyt, Isabelle Boutron, Tammy~C
  Hoffmann, Cynthia~D Mulrow, Larissa Shamseer, Jennifer~M Tetzlaff, Elie~A
  Akl, Sue~E Brennan, et~al.
\newblock The {PRISMA} 2020 statement: An updated guideline for reporting
  systematic reviews.
\newblock \emph{BMJ}, 372:\penalty0 n71, 2021.
\newblock \doi{10.1136/bmj.n71}.

\bibitem[Priem et~al.(2022)Priem, Piwowar, and Orr]{priem2022openalex}
Jason Priem, Heather Piwowar, and Richard Orr.
\newblock {OpenAlex}: A fully-open index of scholarly works, authors, venues,
  institutions, and concepts.
\newblock \emph{arXiv preprint arXiv:2205.01833}, 2022.

\bibitem[Qin et~al.(2023)Qin, Hu, Lin, Chen, Ding, Cui, Zeng, Huang, Xiao, Han,
  Fung, et~al.]{qin2023tool_survey}
Yujia Qin, Shengding Hu, Yankai Lin, Weize Chen, Ning Ding, Ganqu Cui, Zheni
  Zeng, Yufei Huang, Chaojun Xiao, Chi Han, Yi~Ren Fung, et~al.
\newblock Tool learning with foundation models.
\newblock \emph{arXiv preprint arXiv:2304.08354}, 2023.

\bibitem[Schick et~al.(2023)Schick, Dwivedi-Yu, Dess{\`{\i}}, Raileanu, Lomeli,
  Hambro, Zettlemoyer, Cancedda, and Scialom]{schick2023toolformer}
Timo Schick, Jane Dwivedi-Yu, Roberto Dess{\`{\i}}, Roberta Raileanu, Maria
  Lomeli, Eric Hambro, Luke Zettlemoyer, Nicola Cancedda, and Thomas Scialom.
\newblock Toolformer: Language models can teach themselves to use tools.
\newblock In \emph{Advances in Neural Information Processing Systems
  (NeurIPS)}, 2023.
\newblock arXiv:2302.04761.

\bibitem[Schlichtkrull et~al.(2018)Schlichtkrull, Kipf, Bloem, van~den Berg,
  Titov, and Welling]{schlichtkrull2018rgcn}
Michael Schlichtkrull, Thomas~N Kipf, Peter Bloem, Rianne van~den Berg, Ivan
  Titov, and Max Welling.
\newblock Modeling relational data with graph convolutional networks.
\newblock In \emph{European Semantic Web Conference (ESWC)}, pages 593--607,
  2018.
\newblock \doi{10.1007/978-3-319-93417-4_38}.

\bibitem[Tran et~al.(2024)Tran, Gartlehner, Yaacoub, Boutron, Schwingshackl,
  Stadelmaier, Sommer, Alebouyeh, Afach, Meerpohl, and
  Ravaud]{tran2024llm_screening_review}
Viet-Thi Tran, Gerald Gartlehner, Sally Yaacoub, Isabelle Boutron, Lukas
  Schwingshackl, Julia Stadelmaier, Isolde Sommer, Farzaneh Alebouyeh, Sivem
  Afach, Joerg~J Meerpohl, and Philippe Ravaud.
\newblock Sensitivity and specificity of using {GPT}-3.5 turbo models for title
  and abstract screening in systematic reviews and meta-analyses.
\newblock \emph{Annals of Internal Medicine}, 177\penalty0 (6):\penalty0
  791--799, 2024.
\newblock \doi{10.7326/M23-3389}.

\bibitem[{van de Schoot} et~al.(2021){van de Schoot}, de~Bruin, Schram, Zahedi,
  de~Boer, Weijdema, Kramer, Huijts, Hoogerwerf, Ferdinands, Harkema,
  Willemsen, Ma, Fang, Hindriks, Tummers, and
  Oberski]{van_de_schoot2021synergy}
Rens {van de Schoot}, Jonathan de~Bruin, Raoul Schram, Parisa Zahedi, Jan
  de~Boer, Felix Weijdema, Bianca Kramer, Martijn Huijts, Maarten Hoogerwerf,
  Gerbrich Ferdinands, Albert Harkema, Joukje Willemsen, Yongchao Ma, Qixiang
  Fang, Sybren Hindriks, Lars Tummers, and Daniel~L Oberski.
\newblock An open source machine learning framework for efficient and
  transparent systematic reviews.
\newblock \emph{Nature Machine Intelligence}, 3:\penalty0 125--133, 2021.
\newblock \doi{10.1038/s42256-020-00287-7}.

\bibitem[Wang et~al.(2024)Wang, Scells, Koopman, and Zuccon]{wang2024llm_sr}
Shuai Wang, Harrisen Scells, Bevan Koopman, and Guido Zuccon.
\newblock Zero-shot generative large language models for systematic review
  screening automation.
\newblock In \emph{Advances in Information Retrieval (ECIR 2024)}, Lecture
  Notes in Computer Science. Springer, 2024.
\newblock \doi{10.1007/978-3-031-56066-8_32}.
\newblock arXiv:2401.06320.

\bibitem[Wei et~al.(2022)Wei, Wang, Schuurmans, Bosma, Ichter, Xia, Chi, Le,
  and Zhou]{wei2022chain_of_thought}
Jason Wei, Xuezhi Wang, Dale Schuurmans, Maarten Bosma, Brian Ichter, Fei Xia,
  Ed~H Chi, Quoc~V Le, and Denny Zhou.
\newblock Chain-of-thought prompting elicits reasoning in large language
  models.
\newblock In \emph{Advances in Neural Information Processing Systems
  (NeurIPS)}, 2022.

\bibitem[Xiong et~al.(2024)Xiong, Hu, Lu, Li, Fu, He, and
  Hooi]{xiong2024verbalized_uncertainty}
Miao Xiong, Zhiyuan Hu, Xinyang Lu, Yifei Li, Jie Fu, Junxian He, and Bryan
  Hooi.
\newblock Can {LLMs} express their uncertainty? an empirical evaluation of
  confidence elicitation in {LLMs}.
\newblock In \emph{International Conference on Learning Representations
  (ICLR)}, 2024.
\newblock arXiv:2306.13063.

\bibitem[Yao et~al.(2019)Yao, Mao, and Luo]{yao2019textgcn}
Liang Yao, Chengsheng Mao, and Yuan Luo.
\newblock Graph convolutional networks for text classification.
\newblock In \emph{AAAI Conference on Artificial Intelligence}, volume~33,
  pages 7370--7377, 2019.
\newblock \doi{10.1609/aaai.v33i01.33017370}.

\end{thebibliography}

\appendix

\section{Prompt Templates}
\label{app:prompts}

\subsection{System Prompt}

\begin{verbatim}
You are a systematic review screening assistant
for clinical research literature. Your task is to
decide whether a paper should be INCLUDED or EXCLUDED
from a systematic review.

Screening criteria:
[per-dataset criteria, e.g., "Drug class review:
 ACEInhibitors (2544 records, 41 included)"]

Respond in exactly this format:
DECISION: INCLUDE or EXCLUDE
CONFIDENCE: HIGH, MEDIUM, or LOW
REASON: Your 1-2 sentence explanation
\end{verbatim}

\subsection{Baseline User Prompt (Condition 1)}

\begin{verbatim}
Title: [paper title]

Abstract: [paper abstract]
\end{verbatim}

\subsection{Full Spectral Context Block (Condition 2)}

\begin{verbatim}
Title: [paper title]

Abstract: [paper abstract]

--- Spectral Analysis (from a BERT+GCN
    classification model) ---
Model decision: [INCLUDE/EXCLUDE/MAYBE]
Confidence gap: [0.000] (0=uncertain, 1=certain;
                          threshold=0.15)
Model confidence: [0.000]
Paradox detected: [algebraic_radical/
                    categorical_paradox/none]

Note: This analysis is provided as additional context.
You may agree or disagree with the model's assessment
based on your reading of the paper.
\end{verbatim}

\subsection{Decision Only Prompt (Condition 3)}

\begin{verbatim}
Title: [paper title]

Abstract: [paper abstract]

--- Model Analysis ---
A separate BERT+GCN classification model suggests
this paper should be: [INCLUDE/EXCLUDE/MAYBE]
You may agree or disagree with this suggestion
based on your reading.
\end{verbatim}

\subsection{Two-Pass Follow-Up Prompt (Condition 5, Pass 2)}

\begin{verbatim}
You previously assessed this paper with [CONFIDENCE]
confidence. Here is additional analysis from a
classification model:

Title: [paper title]

Abstract: [paper abstract]

--- Spectral Analysis ---
Model decision: [INCLUDE/EXCLUDE/MAYBE]
Confidence gap: [0.000] (0=uncertain, 1=certain;
                          threshold=0.15)
Model confidence: [0.000]
Paradox detected: [algebraic_radical/
                    categorical_paradox/none]

Given this additional context, please reconsider
your decision.
DECISION: INCLUDE or EXCLUDE
CONFIDENCE: HIGH, MEDIUM, or LOW
REASON: Your 1-2 sentence explanation
\end{verbatim}

\section{Hedging Language Patterns}
\label{app:hedging}

The two-pass uncertainty detector uses the following regex pattern
to identify hedging language in LLM responses:

\begin{verbatim}
\b(might|unclear|borderline|uncertain|not sure|
ambiguous|possibly|could be|difficult to determine|
hard to say)\b
\end{verbatim}

Both the explicit confidence check (LOW or MEDIUM) and the hedging
pattern match trigger escalation to a second pass.

\section{Full Per-Dataset Results}
\label{app:full_results}

\begin{table}[htbp]
  \centering
  \caption{Full per-dataset results: mean over 3~seeds $\times$ 5~folds
    for all five conditions and primary metrics.}
  \label{tab:full_results}
  \scriptsize
  \setlength{\tabcolsep}{3pt}
  \begin{tabular}{@{}llccccc@{}}
    \toprule
    \textbf{Dataset} & \textbf{Condition} & \textbf{Precision} & \textbf{Recall} & \textbf{F1} & \textbf{AUC-ROC} & \textbf{WSS@95} \\
    \midrule
    ACEInhibitors          & Baseline        & 0.025 & 0.976 & 0.049 & 0.521 & 0.464 \\
                           & Full spectral   & 0.025 & 0.951 & 0.049 & 0.540 & 0.497 \\
                           & Decision only   & 0.026 & 0.976 & 0.051 & 0.469 & 0.457 \\
                           & MAYBE only      & 0.024 & 0.976 & 0.047 & 0.546 & 0.466 \\
                           & Two-pass        & 0.026 & 0.959 & 0.051 & 0.488 & 0.446 \\
    \addlinespace
    ADHD                   & Baseline        & 0.118 & 0.840 & 0.207 & 0.457 & 0.781 \\
                           & Full spectral   & 0.140 & 0.890 & 0.242 & 0.490 & 0.880 \\
                           & Decision only   & 0.126 & 0.840 & 0.219 & 0.479 & 0.780 \\
                           & MAYBE only      & 0.117 & 0.840 & 0.205 & 0.480 & 0.779 \\
                           & Two-pass        & 0.130 & 0.824 & 0.224 & 0.481 & 0.789 \\
    \addlinespace
    Antihistamines         & Baseline        & 0.069 & 0.957 & 0.129 & 0.527 & 0.544 \\
                           & Full spectral   & 0.073 & 0.943 & 0.135 & 0.540 & 0.597 \\
                           & Decision only   & 0.073 & 0.943 & 0.135 & 0.489 & 0.522 \\
                           & MAYBE only      & 0.070 & 0.970 & 0.130 & 0.525 & 0.545 \\
                           & Two-pass        & 0.075 & 0.943 & 0.138 & 0.547 & 0.581 \\
    \addlinespace
    NSAIDs                 & Baseline        & 0.120 & 1.000 & 0.214 & 0.602 & 0.361 \\
                           & Full spectral   & 0.126 & 1.000 & 0.224 & 0.631 & 0.446 \\
                           & Decision only   & 0.137 & 1.000 & 0.241 & 0.552 & 0.409 \\
                           & MAYBE only      & 0.119 & 1.000 & 0.213 & 0.607 & 0.375 \\
                           & Two-pass        & 0.143 & 1.000 & 0.251 & 0.548 & 0.400 \\
    \addlinespace
    OralHypoglycemics      & Baseline        & 0.317 & 0.850 & 0.462 & 0.480 & 0.116 \\
                           & Full spectral   & 0.326 & 0.889 & 0.476 & 0.507 & 0.164 \\
                           & Decision only   & 0.331 & 0.880 & 0.481 & 0.486 & 0.131 \\
                           & MAYBE only      & 0.315 & 0.898 & 0.467 & 0.510 & 0.145 \\
                           & Two-pass        & 0.322 & 0.890 & 0.473 & 0.489 & 0.159 \\
    \addlinespace
    ProtonPumpInhibitors   & Baseline        & 0.054 & 0.947 & 0.103 & 0.582 & 0.434 \\
                           & Full spectral   & 0.061 & 0.921 & 0.115 & 0.554 & 0.473 \\
                           & Decision only   & 0.062 & 0.941 & 0.115 & 0.556 & 0.474 \\
                           & MAYBE only      & 0.055 & 0.954 & 0.104 & 0.576 & 0.431 \\
                           & Two-pass        & 0.064 & 0.934 & 0.119 & 0.527 & 0.440 \\
    \addlinespace
    Triptans               & Baseline        & 0.051 & 0.897 & 0.096 & 0.528 & 0.560 \\
                           & Full spectral   & 0.055 & 0.897 & 0.103 & 0.480 & 0.512 \\
                           & Decision only   & 0.055 & 0.897 & 0.103 & 0.468 & 0.499 \\
                           & MAYBE only      & 0.051 & 0.897 & 0.096 & 0.520 & 0.521 \\
                           & Two-pass        & 0.054 & 0.858 & 0.102 & 0.538 & 0.527 \\
    \addlinespace
    UrinaryIncontinence    & Baseline        & 0.230 & 0.845 & 0.360 & 0.513 & 0.513 \\
                           & Full spectral   & 0.288 & 0.872 & 0.429 & 0.490 & 0.604 \\
                           & Decision only   & 0.275 & 0.787 & 0.405 & 0.495 & 0.463 \\
                           & MAYBE only      & 0.229 & 0.859 & 0.361 & 0.519 & 0.512 \\
                           & Two-pass        & 0.278 & 0.809 & 0.412 & 0.495 & 0.495 \\
    \addlinespace
    \bottomrule
  \end{tabular}
\end{table}

\end{document}